\documentclass[sigconf]{acmart}

\AtBeginDocument{%
  }

\begin{document}

\title{MVQA-68K: A Multi-dimensional and Causally-annotated Dataset with Quality Interpretability for Video Assessment}

\author{Yanyun Pu}
\orcid{0009-0003-7120-2777}
\affiliation{%
	\institution{Huawei Technologies Co.}
	\city{Hangzhou}
	\state{Zhejiang}
	\country{China}
}
\email{puyanyun@huawei.com}

\author{Kehan Li}
\authornote{Corresponding author}
\orcid{0000-0002-5318-4806}
\affiliation{%
	\institution{Huawei Technologies Co.}
	\city{HangZhou}
	\state{Zhejiang}
	\country{China}}
\email{likehan1@hauwei.com}

\author{Zeyi Huang}
\orcid{0009-0009-5897-2172}
\affiliation{%
	\institution{Huawei Technologies Co.}
	\city{HangZhou}
	\state{Zhejiang}
	\country{China}
}
\email{huangzeyi2@huawei.com}

\author{Zhijie Zhong}
\orcid{0000-0003-0203-8419}
\affiliation{%
	\institution{South China University of Technology}
	\city{GuangZhou}
	\state{Guangdong}
	\country{China}}
\email{csemor@mail.scut.edu.cn}

\author{Kaixiang Yang}
\authornotemark[1]
\orcid{0000-0003-2180-2101}
\affiliation{%
	\institution{South China University of Technology}
	\city{GuangZhou}
	\state{Guangdong}
	\country{China}}
\email{yangkx@scut.edu.cn}


\renewcommand{\shortauthors}{Yanyun Pu, Kehan Li, Zeyi Huang, Zhijie Zhong, and Kaixiang Yang}

\begin{abstract}

With the rapid advancement of video generation models such as Sora, video quality assessment (VQA) is becoming increasingly crucial for selecting high-quality videos from large-scale datasets used in pre-training. Traditional VQA methods, typically producing single numerical scores, often lack comprehensiveness and interpretability. To address these challenges, we introduce \textbf{MVQA-68K}, a novel multi-dimensional VQA dataset comprising over 68,000 carefully annotated videos, covering seven essential quality dimensions: overall aesthetics, camera movement, dynamic degree, texture detail, composition, visual quality, and factual consistency. Each annotation includes detailed chain-of-thought reasoning to facilitate interpretability and comprehensive understanding. Extensive experiments demonstrate that MVQA-68K significantly enhances the performance of various multimodal large language models (MLLMs) on the VQA task, achieving state-of-the-art results not only on our internal test set (Fig.~\ref{fig1}) but also on public benchmarks including LSVQ-test, LSVQ-1080p, and LIVE-VQC. Meantime,  incorporating explicit reasoning  process during VQA training substantially boosts the zero-shot generalization. Code and dataset will be available at \url{github: https://github.com/Controller01-ai/MVQA-68K}

\end{abstract}



\maketitle

\begin{figure}[htbp]
\centering
\includegraphics[width=0.45\linewidth]{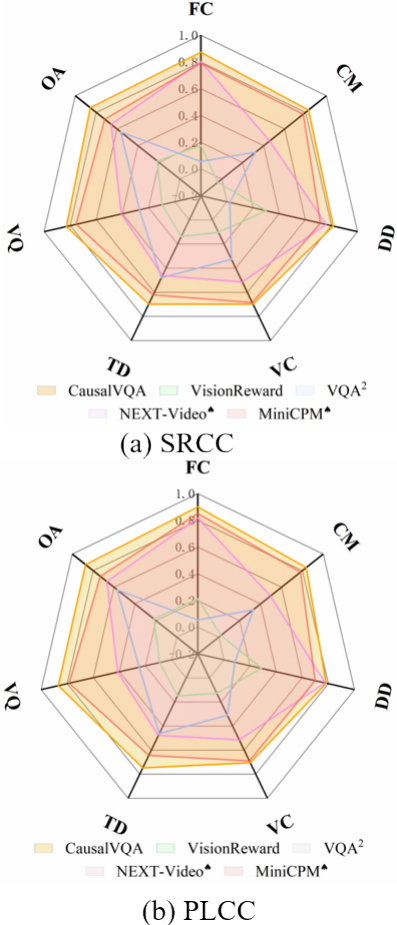}
\caption{The SRCC (a) and PLCC (b) performance comparison of CausalVQA, VisionReward, VQA$^{2}$, NEXT-Video$^{\spadesuit}$, and MiniCPM$^{\spadesuit}$ on the MVQA-test dataset. VisionReward and VQA$^{2}$ were evaluated in a zero-shot setting, while the remaining models were fine-tuned on the MVQA-68K training set.}\label{fig1}
\end{figure}
\section{Introduction}
VQA is pivotal in evaluating the suitability and effectiveness of video content \cite{bib_82}, particularly in the era of large-scale AI-generated content (AIGC) models, where high-quality data directly influence model capabilities \cite{bib_1_3, bib_1_4}. However, obtaining high-quality videos from vast User-Generated Content (UGC) and Professionally Generated Content (PGC) remains challenging, given the diverse range of content styles, aesthetic standards, and shooting techniques\cite{bib_1_1}. These complexities pose significant challenges in designing VQA models, necessitating a thorough understanding of  video data across multiple dimensions, in order to facilitate the development of a more robust and generalizable AIGC model.
\begin{table*}[ht]
	\begin{center}
		\caption{The latest MLLMs' performance on different publicly available VQA datasets. We require the MLLMs to generate a specific score.
		}\label{table_0}
		\includegraphics[width=0.9\linewidth]{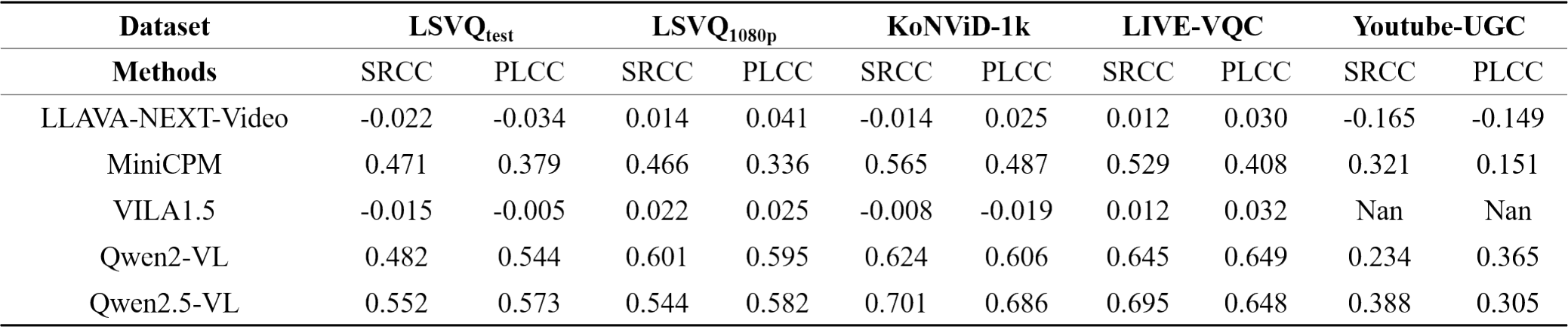}\\
	\end{center}
\end{table*}

Unlike static images, real-world videos quality assement are especially diffucult due to complex camera movements, varying degrees of dynamic content, intricate texture details, and subjective aesthetic preferences. These dimensions not only influence perceptual quality but also directly impact the performance of AIGC models.  As shown in Fig.\ref{fig3}, these dimensions are highly interrelated, posing challenges for comprehensive and accurate quality assessment. Existing approaches typically rely solely on numerical quality scores, failing to capture the nuanced interplay across dimensions \cite{bib_6, bib_7, bib_8, bib_9, bib_10}. Moreover, the subjective nature of scoring leads to inconsistent annotations, limiting model generalization capability. Beyond, the lack of reasoning process annotation also makes it hard for VQA model to essentially understand quality shortcomings in video data, leading to poor interpretability. Therefore, a systematic, interpretable, and multidimensional framework is urgently needed to facilitate precise identification of quality issues and improve AIGC model performance.\cite{bib_11, bib_12, bib_13}. 

Recently, multi-modal large language models (MLLMs) offer new possibilities for addressing the complexities of video quality assessment. By combining linguistic reasoning with visual understanding, MLLMs can evaluate videos not only on fundamental dimensions but also on more abstract factors such as overall aesthetics \cite{bib_16, bib_17}. However, directly applying existing MLLMs to VQA faces several limitations. First, most MLLMs offer coarse-grained evaluations, lacking sensitivity to nuanced multidimensional quality factors. Second, although some UGC datasets include multiple quality dimensions, they typically focus only on basic aspects like blur or artifacts, overlooking more complex and subjective quality dimensions \cite{bib_39, bib_40}. Third, aesthetic evaluations remain highly subjective, causing MLLMs to overly rely on superficial label alignment, thus impairing generalization across diverse content types \cite{bib_18, bib_19, bib_20}. As illustrated in Tab.~\ref{table_0}, even state-of-the-art MLLMs (e.g., Qwen2.5-VL) achieve relatively low correlation metrics (SRCC, PLCC) on mainstream VQA benchmarks, underscoring the need for a more comprehensive and nuanced dataset.

\begin{figure*}[ht]
\centering
\includegraphics[width=0.9\linewidth]{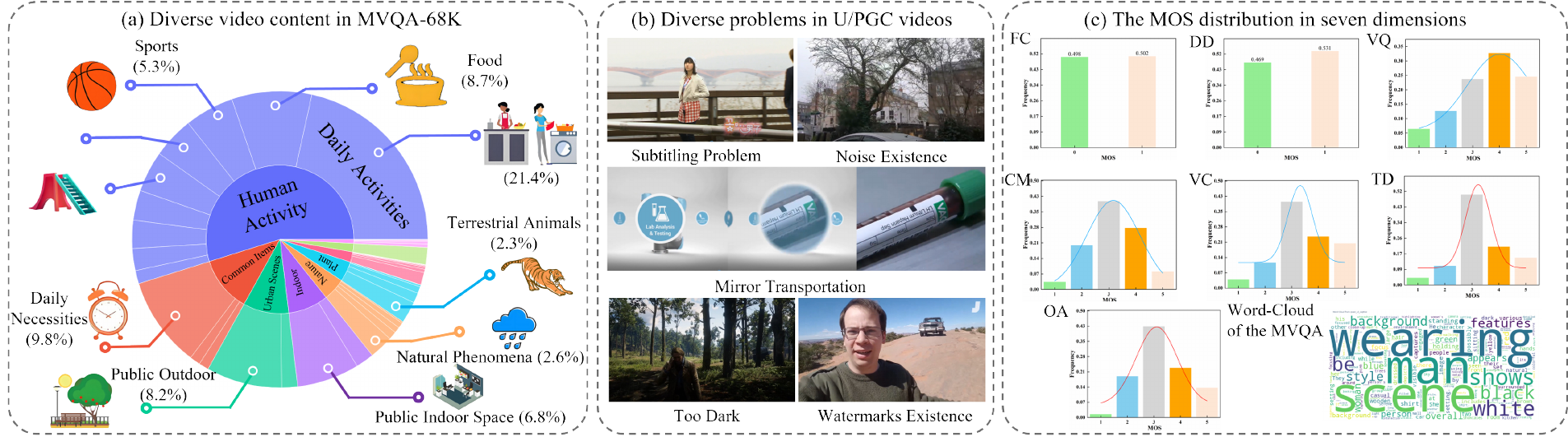}
\caption{We present MVQA-68K and the corresponding model CausalVQA for fine-grained video quality assessment. As illustrated in subfigures (a), (b), and (c), the dataset contains diverse video content exhibiting a wide range of quality issues. This highlights the need for multi-dimensional quality labeling beyond overall scores to support downstream tasks such as video enhancement and recommendation. To address data imbalance, we apply resampling, and the resulting MOS distributions across key dimensions are visualized in (c).}\label{fig2}
\end{figure*}

To address these issues, we introduce the MVQA-68K dataset, a novel multidimensional video quality assessment benchmark comprising 38,254 carefully curated videos annotated across seven critical dimensions: overall aesthetics, camera movement, dynamic degree, texture detail, video composition, visual quality, and factual consistency. As shown in Fig.~\ref{fig2}, our dataset explicitly includes high-end aesthetic videos to overcome limitations inherent in existing benchmarks, which predominantly consist of relatively lower-quality videos \cite{bib_21,bib_22,bib_23,bib_24,bib_25}. Additionally, MVQA-68K incorporates detailed causal rationale annotations generated with both human experts and LLM, totaling over 68,000 expert-verified data points. To further enhance assessment reliability, we introduce a multi-prompt ensemble (MPE) strategy \cite{bib_26}, leveraging diverse semantic prompts to achieve more stable and precise quality evaluations. Specifically, diverse prompts expand the semantic coverage and helps mitigate ambiguity caused by single expressions. Extensive experiments on multiple benchmarks demonstrated that models fine-tuned on MVQA-68K achieve state-of-the-art or near state-of-the-art performance across various public benchmarks, significantly improving zero-shot generalization capabilities. 

Our contributions are summarized as follows:
\begin{enumerate}
	\item We construct \textbf{MVQA-68K}, a large-scale VQA benchmark consisting of 38,254 diverse videos and 68,783 multi-dimensional annotations, uniquely incorporating causal reasoning explanations to enable interpretable quality assessment.
	\item Leveraging MVQA-68K, we develop \textbf{CausalVQA}, a novel model based on Qwen2-VL-7B, achieving superior performance compared to existing state-of-the-art methods across multiple VQA benchmarks.
	\item We propose \textbf{InsightVQA}, a human-aligned interpretability framework, offering transparent, causal insights into model predictions, thereby advancing interpretable standards for video quality assessment.
\end{enumerate}

\section{Realted Work}
\subsection{Datasets and Benchmarks for VQA}
Many video quality assessment datasets have been established to analyse human perceptual responses to video content. Early datasets focused on synthetic distortions, relied on a limited set of original videos, and manually introduced degradation types \cite{bib_28, bib_29, bib_30}. However, with the increasing popularity of UGC, researchers have turned to real-world distortions in realistic videos. Several UGC-VQA datasets \cite{bib_31, bib_32, bib_33, bib_34} have been developed to evaluate in-camera and real-world distortions, reflecting the diversity of quality differences in UGC. Other datasets \cite{bib_35, bib_36, bib_37} combine synthetic and real distortions, aiming to bridge the gap between controlled experiments and real-world video assessment. Moreover, considering that most UGC-VQA datasets primarily obtain videos from general media platforms (e.g., YouTube), recent works such as KVQ \cite{bib_38} introduce datasets especially for short videos, addressing the unique quality challenges of this fast-growing video format. Despite these advances, existing datasets often lack high-end aesthetic considerations \cite{bib_39, bib_40, bib_41} and fail to provide fine-grained and interpretable quality assessments across multiple dimensions.

The rapid development of LLMs \cite{bib_42,bib_43,bib_44} has showed their exceptional abilities in various image-based benchmarks, achieving breakthroughs in tasks such as image captioning and visual question answering \cite{bib_45, bib_46,bib_47,bib_48}. Models such as CLIP \cite{bib_49} and ALIGN \cite{bib_50} have set new standards by leveraging large-scale paired datasets and transformer architectures. As research progresses, the focus has expanded from images to videos, which present unique challenges due to additional temporal dynamics and motion consistency. Early efforts, such as VideoBERT \cite{bib_51} and ViLT \cite{bib_52}, demonstrated their potential but often lacked diversity in video types and failed to capture the nuanced temporal attributes of real-world videos. Recent benchmarks have shifted toward more comprehensive evaluations \cite{bib_53, bib_54}, but they remain largely centered on high-level semantics \cite{bib_55, bib_56}. For instance, VideoScore \cite{bib_2} assumes that UGC videos, as real-world content, inherently achieve high quality across all evaluation dimensions. However, this assumption overlooks the fact that many real-world videos suffer from poor quality, including issues such as unstable camera movement, compression artifacts, and lack of aesthetic coherence.  

Additionally, recent works like Q-Bench-Video \cite{bib_57} and Q-Instruct \cite{bib_58} have explored visual scoring with LMMs, introducing strategies such as binary softmax for quality prediction and fine-tuning with low-level queries. These approaches highlight the potential of LLMs in visual scoring \cite{bib_59} but are often limited by simplified methodologies and insufficient datasets. 
\begin{figure*}[htbp]
\centering
\includegraphics[width=0.92\textwidth]{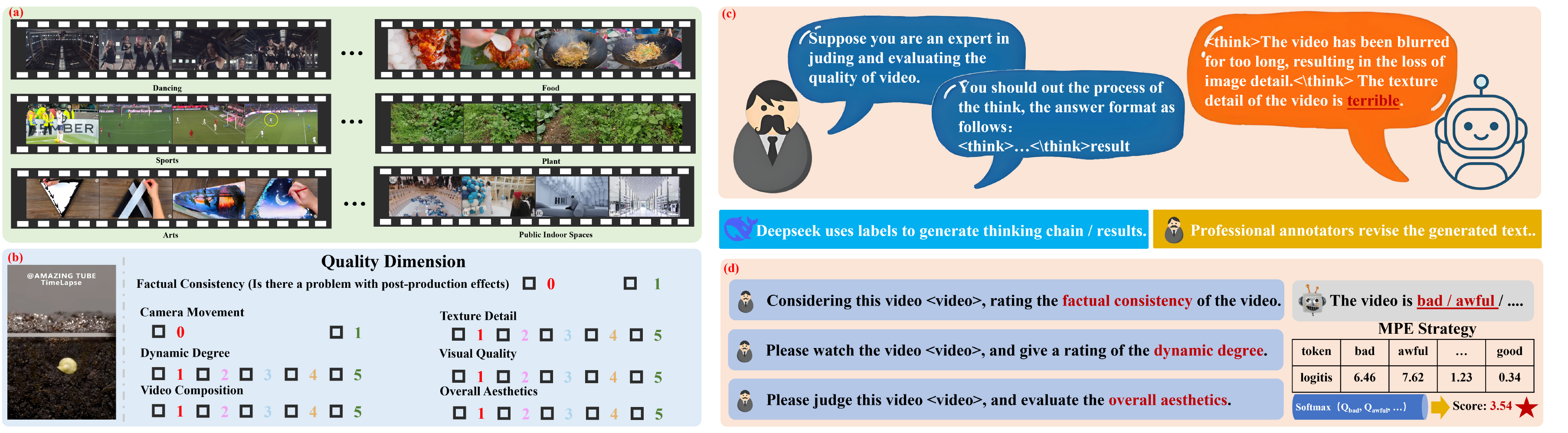}
\caption{Overview of the content and construction process of MVQA-68K. (a) Sample videos in the dataset, which contain a variety of common UGC videos. (b) The subjective data annotation process, in which multiple people score. (c) The relevant question-answer pairs generated by Deepseek using the scoring reasons of each dimension and modified by human annotators, which guide the thinking of MLLM. (d) The probability of the token is calculated between multiple prompts.}\label{fig3}
\end{figure*}

\subsection{Algorithom for VQA}
\subsubsection{Knowledge-Driven VQA Models}
Unlike tasks such as action recognition \cite{bib_60} or object detection \cite{bib_61}, VQA focuses on perceptual attributes like motion consistency, texture detail, and aesthetic appeal, which are vital for generative video applications \cite{bib_62}. Early VQA methods primarily relied on knowledge-driven techniques, using handcrafted features and statistical metrics such as PSNR \cite{bib_63} and SSIM \cite{bib_64} to capture low-level distortions. To better align with human perception, models like NIQE \cite{bib_65} and BRISQUE \cite{bib_66} introduced natural scene statistics (NSS), while V-BLIINDS \cite{bib_67} and VIIDEO \cite{bib_68} extended these to the temporal domain.

Subsequent approaches, including TLVQM \cite{bib_69} and VIDEVAL \cite{bib_70}, integrated multi-level or diverse handcrafted features to enhance robustness. Perceptual fusion metrics such as VMAF \cite{bib_71,bib_72} further improved correlation with human Mean Opinion Scores (MOS) \cite{bib_73,bib_74}. However, these models often underperform on in-the-wild UGC videos due to their limited capacity to model complex content variations and high-level perceptual cues. 

\subsubsection{Data-driven VQA Models}
With the rise of deep learning, data-driven VQA models have emerged as effective alternatives to traditional handcrafted approaches by learning quality-aware representations directly from large-scale video datasets. These end-to-end frameworks better align with human perception and offer improved robustness in out-of-distribution scenarios. A variety of architectures have been proposed for spatial and temporal quality modeling. For example, VSFA \cite{bib_75} employs deep feature embeddings, DisCoVQA \cite{bib_76} introduces contrastive learning for distortion awareness, and SimpleVQA \cite{bib_77} combines high-resolution keyframes with low-resolution motion features for efficient inference. RAPIQUE \cite{bib_78} further integrates handcrafted and learned features to enhance generalization.

To better capture temporal dependencies, models based on recurrent units \cite{bib_79}, hybrid CNN-RNN structures \cite{bib_81}, and transformer-based designs such as FAST-VQA \cite{bib_82} have been developed. The latter utilizes spatially spliced and temporally aligned patches via a modified Swin Transformer \cite{bib_83}, improving long-range modeling. Despite their progress, data-driven VQA models often rely heavily on large annotated datasets and still struggle with generalization to unseen or complex UGC scenarios.

\subsubsection{MLLMs for VQA}
The emergence of multimodal large language models (MLLMs) has opened new avenues for VQA by unifying visual and linguistic understanding within a single framework \cite{bib_84, bib_40, bib_53, bib_11}. Unlike traditional methods based on handcrafted features or deep networks, MLLMs enable more interpretable and context-aware quality evaluation. Recent efforts, such as Q-ALIGN \cite{bib_59}, integrate frame-level visual analysis with textual reasoning to bridge video and image quality assessment. Other approaches adopt token-based scoring strategies, where quality labels like “good” or “poor” are mapped using softmax outputs \cite{bib_11}. However, these methods often suffer from limited visual-text alignment, particularly in modeling temporal dynamics and fine-grained distortions. To address this, Q-Instruct \cite{bib_58} introduces instruction-tuned datasets focused on low-level visual quality cues, enhancing MLLMs’ sensitivity to perceptual attributes. However, current VQA methods based on MLLM still face challenges in accurately capturing complex spatiotemporal quality factors, and their generalizability remains to be tested.

\section{MVQA-68K dataset}
In this section, we present \textbf{MVQA-68K}, a large-scale and fine-grained video quality assessment dataset containing 38,254 videos and 825,396 human annotations across diverse real-world scenarios. It covers multiple quality dimensions with rich reasoning-based labels, addressing the limitations of existing coarse-grained UGC datasets. By incorporating detailed justifications, MVQA-68K facilitates deeper chain-of-thought (CoT) modeling in MLLMs.

\subsection{Video Collection}
MVQA-68K is built from high-quality public datasets such as Panda-70M \cite{panda-70M} and Koala-36M \cite{koala}, ensuring diverse and representative video content for quality assessment. To improve data reliability, we apply TransNetV2 \cite{transnetv2} for scene transition detection and PyScene-Detect to remove static or low-motion clips. The dataset covers a wide range of real-world scenarios, including human activities, urban scenes, animals, plants, and indoor environments (see Fig.~\ref{fig2}). This diversity supports the development of models capable of evaluating both low-level distortions and high-level perceptual attributes across various video domains.

\subsection{Annotation Pipeline}
\subsubsection{Evaluation Dimensions}
Inspired by prior works such as VBench \cite{bib_17}, EvalCrafter \cite{bib_85}, FETV \cite{bib_86}, and VideoFeedback \cite{bib_2}, our dataset defines seven key dimensions, as summarized in Fig.~\ref{fig3}. Texture Detail focuses on clarity, edge sharpness, and material fidelity, while Visual Quality evaluates degradations like blur, noise, or flicker. Dynamic Degree captures temporal consistency, and Camera Movement assesses the naturalness and smoothness of motion such as panning or zooming. Video Composition measures structural balance and subject framing. At a higher perceptual level, Factual Consistency checks content authenticity, primarily intended for identifying content without post-production editing such as watermark, and Overall Aesthetics reflects subjective appeal through factors like lighting and color harmony. Each dimension is supported by detailed annotation guidelines to ensure consistent and objective evaluations among annotators. 

\subsubsection{Annotation Protocol}
To ensure annotation accuracy, 14 trained annotators independently evaluated each video across the seven dimensions. Each dimension was rated by 6 to 7 annotators, and final scores were calculated by discarding the highest and lowest values and averaging the rest. Annotators received structured training and underwent quality checks throughout the process to maintain consistency. We adopted a binary scoring system for factual consistency and dynamic degree: 0 indicates non-factual or static content, and 1 indicates factual or dynamic content. For the other dimensions, we used a five-point scale (1–5), as shown in Fig.~\ref{fig3}(b), allowing for fine-grained quality differentiation.

To reduce label bias, binary attributes were balanced between classes, while five-level scores were adjusted to follow a natural, approximately normal distribution, as shown in Fig.~\ref{fig3}(c). This reflects the real-world distribution of UGC video quality, where extreme scores are less common. Our annotation strategy ensures the dataset offers structured, high-quality, and representative labels for training and evaluation.
\begin{figure*}[htbp]
\centering
\includegraphics[width=0.9\textwidth]{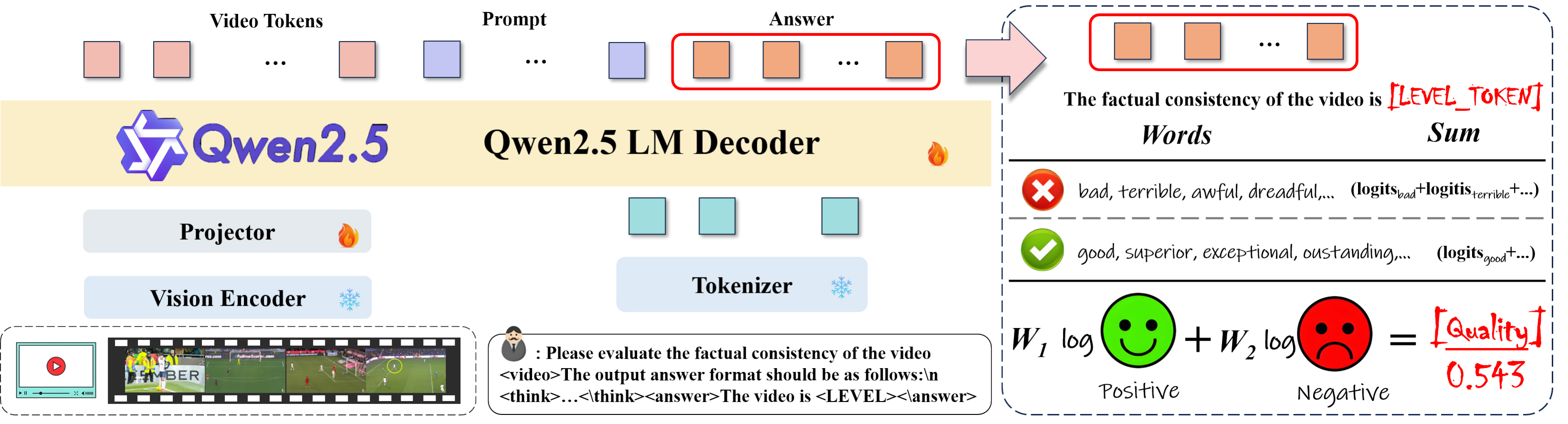}
\caption{Model structure of Qwen2.5-VL with MVQA-68K fine-tuning.}\label{fig4}
\end{figure*}
\subsection{QA Pairs for Quality Score}
To better align with the language understanding capabilities of MLLMs, we convert numerical MOS scores into descriptive labels that reflect human-like reasoning. Specifically, scores from 1 to 5 are mapped to categorical labels: "bad", "poor", "fair", "good", and "excellent". To further improve robustness, we adopt a Multi-Prompt Ensemble (MPE) strategy by enriching each label with semantically related expressions (e.g., "bad" includes "terrible", "awful", "dreadful"). This lexical diversity mitigates ambiguity and improves generalization, enabling MLLMs to produce more stable and accurate quality assessments. 

\subsection{QA Pairs for Video Understanding}
This annotation phase introduces explanation-based QA pairs to enhance video understanding. Annotators provide brief justifications alongside quality scores for each dimension. These human-written explanations are then expanded using Deepseek into detailed, coherent rationales that reflect how human evaluators perceive video quality. For instance, a short note like “some areas lack detail” may be expanded to a full explanation highlighting motion blur and resolution loss. Each generated explanation is manually verified to ensure factual accuracy and alignment with the video content. This human-in-the-loop process produces high-quality QA pairs that pair natural language reasoning with semantic labels.

\section{Methods}
\subsection{Problem Formulation}
Given a video $X$, our goal is to estimate its perceptual quality across multiple predefined dimensions using a MLLM. Unlike traditional VQA approaches that predict numerical MOS values, we generate descriptive quality labels (e.g., "poor", "good", "excellent") that better align with human perception.

The video is represented as a frame sequence $\{x_{i}\}_{i = 0}^{N - 1}$, where $x_{i} \in \mathbb{R}^{H \times W \times 3}$, and $N$ is the number of frames. We uniformly sample $M \ll N$ frames to obtain a compact representation $X^{\prime}=\{x_{j}^{\prime}\}_{j=0}^{M - 1}$, which is fed into the MLLM without any external text prompt.
The quality assessment model is defined as: \begin{equation} \hat{Q}=F\left(X^{\prime}\right) \end{equation} where $F$ is the MLLM and $\hat{Q} \in \mathcal{L}$ is the predicted label selected from the set $\mathcal{L} = {\text{bad}, \text{poor}, \text{fair}, \text{good}, \text{excellent}}$ or their lexical variants, as used in our multi-prompt ensemble strategy. 

\subsection{Model Structure}
We adopt the pre-trained vision-language model Qwen-VL-2.5-7B for multi-dimensional video quality assessment, as illustrated in Fig.~\ref{fig4}. Without modifying its architecture, we fine-tune the model on MVQA-68K to align it with human-perceived video quality. To further improve its reasoning ability, we apply a causal prompt engineering strategy, where dimension-specific instructions are used to guide the model toward logically grounded and human-aligned predictions. This allows the model to not only generate descriptive quality labels but also explain the rationale behind its judgments.

\subsection{SemLogitMPE}
To mitigate ambiguity in prompt-based evaluation, we propose SemLogitMPE, a multi-prompt ensemble strategy that expands each quality label (e.g., “poor”) into a set of semantically related expressions (e.g., “inferior,” “mediocre,” “substandard”). Each prompt is independently passed to the model, and the corresponding pre-softmax logits are aggregated as:
\begin{equation}
    \hat{\mathcal{P}}_{l o g}\left(\mathcal{W}_{\alpha}\right)= \sum \mathcal{P}_{l o g}\left(\mathcal{W}_{i}\right)
\end{equation}
where $\alpha \in\{bad, poor, fair, good, excellent\}$, $i$ indexes synonym variants.This aggregation captures a broader semantic space and stabilizes predictions by reducing reliance on individual phrasings. 

\section{Experiment and Result}
\subsection{Experimental Setting}
\subsubsection{Implementation Details}
We fine-tune the Qwen-VL-2.5-7B model while keeping the vision encoder frozen. Training is performed on 8 GPUs (80 GB each) using bf16 precision and FlashAttention for improved efficiency. The global batch size is 64, with learning rates set to 1e-5 for the LLM backbone and 2e-6 for the video projection module. To avoid overfitting, we train for one epoch. During preprocessing, frames are uniformly sampled at 1 fps and resized to $128\times28\times28$. 
\begin{table*}[ht]
  \begin{center}
  \caption{Performance comparison of traditional VQA methods, DNN-based VQA, generic MLLMs and CausalVQA on the seven benchmark dimensions of the MVQ database. ‘NaN’ indicates that no valid result was output.
}\label{table_2}
\includegraphics[width=0.9\linewidth]{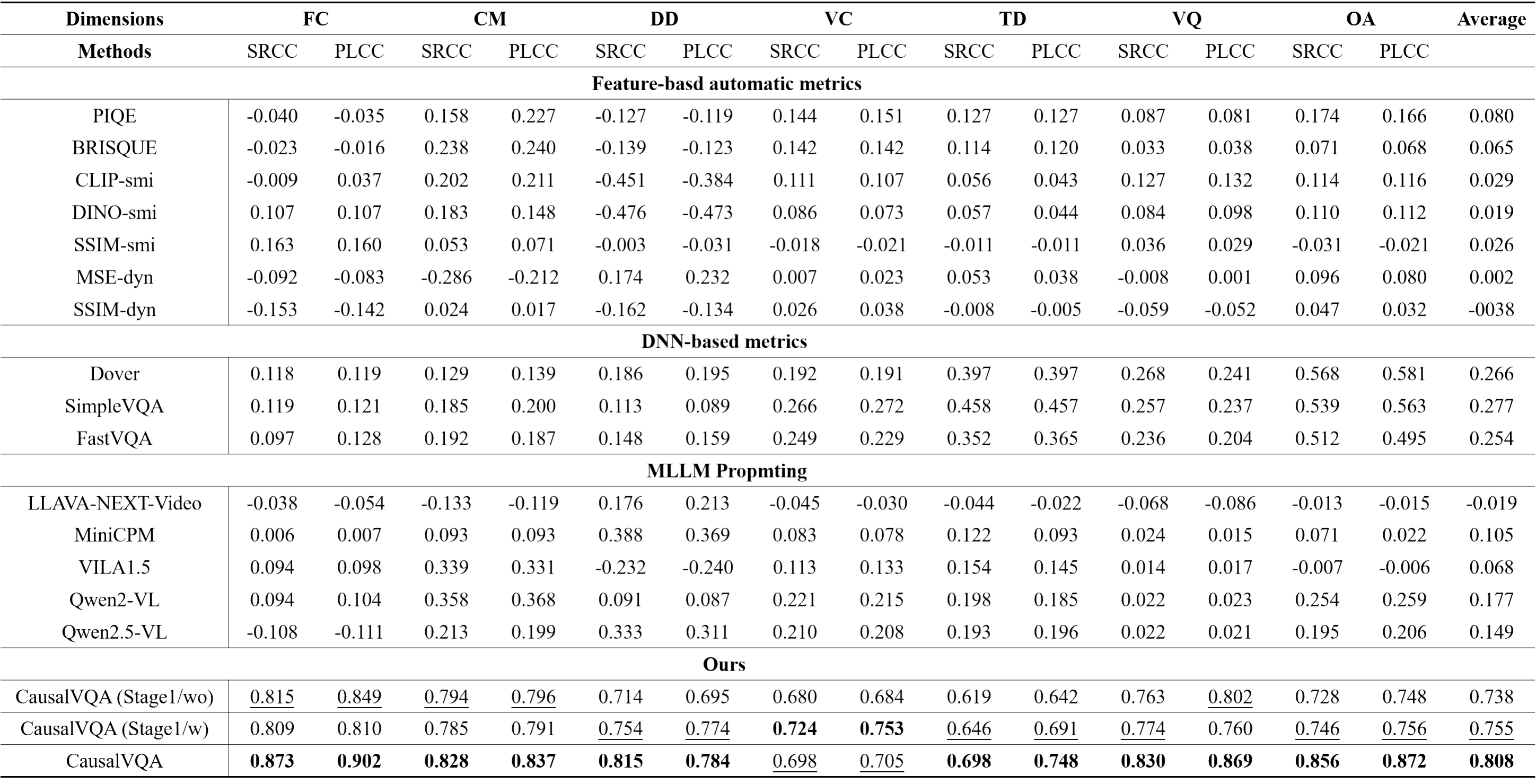}\\
  \end{center}
\end{table*}
\subsubsection{Video Test Dataset}
To evaluate the generalization ability of our fine-tuned MVQA model, we conduct cross-dataset testing on five public VQA benchmarks: YouTube-UGC \cite{bib_25}, LSVQ-test \cite{bib_21}, LSVQ-1080p \cite{bib_21}, LIVE-VQC \cite{bib_24}, and KoNViD-1k \cite{bib_22}. These datasets cover diverse real-world UGC scenarios with varied content types. We use the original mean opinion scores (MOS) from each dataset and focus solely on overall quality prediction. No additional annotations are added, ensuring a consistent and rigorous cross-dataset evaluation setting.
\begin{table*}[ht]
	\begin{center}
		\caption{Performance comparison of four specialized video quality assessment models on the MVQA-test dataset across seven evaluation dimensions. Results are reported in terms of SRCC and PLCC.
		}\label{table_3}
		\includegraphics[width=0.85\linewidth]{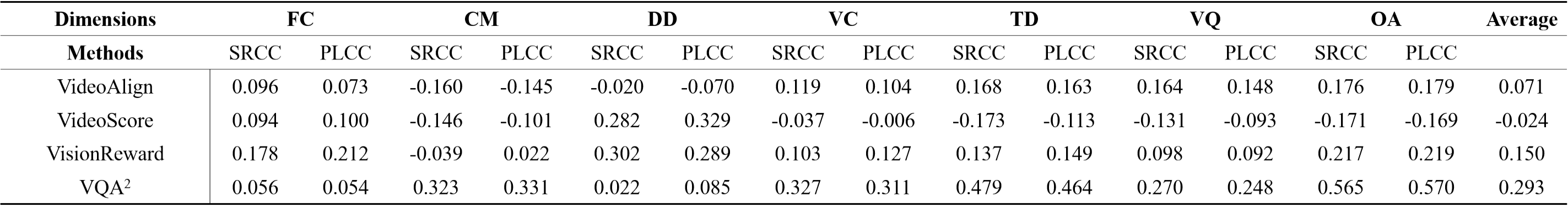}\\
	\end{center}
\end{table*}
\begin{table*}[ht]
	\begin{center}
		\caption{Performance comparison of three CausalVQA variants and four professional VQA models on five public video quality datasets.
		}\label{table_4}
		\includegraphics[width=0.85\linewidth]{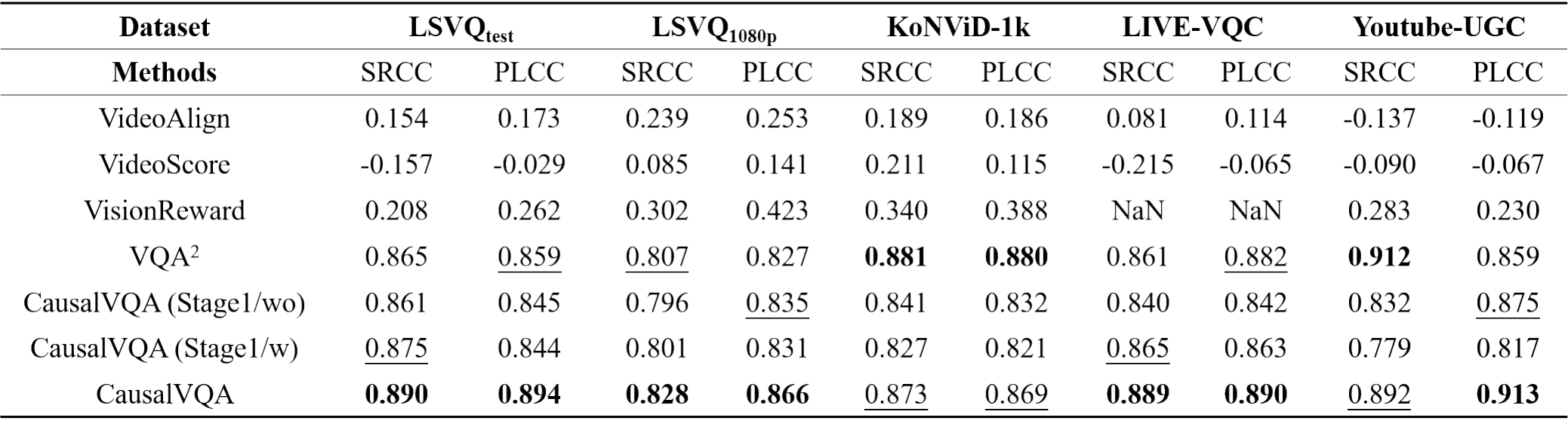}\\
	\end{center}
\end{table*}

\subsubsection{Comparison Models}
We compare our method against a wide range of baselines, including traditional feature-based metrics, deep learning-based VQA models, general-purpose multimodal models, and specialized video quality assessment frameworks. Evaluations are conducted on both the MVQA-68K test set and several public benchmarks. Deep learning-based models include Dover \cite{dover}, SimpleVQA \cite{bib_77}, and FAST-VQA \cite{bib_82}, which represent recent progress in blind VQA. We also benchmark several general-purpose MLLMs. Lastly, we evaluate specialized VQA models such as VideoAlign \cite{bib_3}, VQA$^{2}$ \cite{VQA2}, which are designed for fine-grained quality evaluation or video generation feedback.

\subsection{Performance Comparison}
\subsubsection{Results on the MVQA-test with Generic Model}
We evaluate our fine-tuned CausalVQA model (based on Qwen2.5-VL-7B) and multiple baselines on the MVQA-test set, as shown in Tab.~\ref{table_2}. Based on the MVQA-68K dataset, we derive three variants of the CausalVQA model. Specifically: (1) CausalVQA (Stage1/wo), which excludes both the MPE strategy and causal inference; (2) CausalVQA (Stage1/w), which incorporates the MPE strategy but still excludes causal inference; and (3) the full version of CausalVQA, which leverages both the MPE strategy and causal inference during fine-tuning. It is worth noting that, unless explicitly stated otherwise, all reported results correspond to zero-shot evaluations. Traditional feature-based metrics show inconsistent performance and fail to capture perceptual quality across dimensions. Deep learning-based models perform more stably, particularly on texture details and overall aesthetics, but still struggle with dimensions like factual consistency and dynamic degree. General-purpose MLLMs show moderate results but lack fine-grained perceptual alignment. In contrast, our CausalVQA model consistently outperforms all baselines across the seven dimensions, with notable gains in subjective aspects like overall aesthetics and factual consistency. 
\subsubsection{Results on the MVQA-test with Professional Model}

We evaluate four professional VQA models on the MVQA-test. Although these models perform well on their respective training datasets, their results reveal limited generalization when applied to MVQA-test. As shown in Tab.~\ref{table_3}, correlations with human annotations drop significantly. Among them, VQA$^{2}$ achieves the highest average correlation, yet still underperforms in capturing fine-grained perceptual quality.

\subsubsection{Results on the Existing Public Benchmark}
We compare the three CausalVQA variants with four professional model on five public VQA datasets: YouTube-UGC, LIVE-VQC, LSVQ-test, LSVQ-1080p, and KoNViD-1k. Tab.~\ref{table_4} reports the results of each method on each benchmark. CausalVQA  achieves the highest overall performance and the ablation variants. In fact, the full model delivers consistently strong correlations on every dataset (each SRCC > 0.82), despite being evaluated on these datasets for the first time. Among the baselines, VQA2 is the strongest competitor, but CausalVQA (final) still surpasses it on most benchmarks and in the overall mean. Notably,  We observe that introducing the multi-prompt ensemble (Stage1/w vs. Stage1/wo) yields only modest and dataset-dependent gains: e.g., +0.025 SRCC on LIVE-VQC but a slight drop on YouTube-UGC, resulting in similar average performance. In contrast, incorporating causal reasoning in the final model boosts accuracy across the board , raising the mean SRCC by ~4–5 points relative to the Stage1 variants and notably improving the more challenging datasets.
\subsection{Data Portfolio Strategy}
This subsection investigates how training data composition affects video quality assessment across dimensions. We compare single-dimension training, where models are trained on one quality aspect, with multi-dimension training using QA pairs from all dimensions. Experiments are conducted under two settings: CausalVQA (Stage1/wo), without multi-prompt ensemble (MPE), and CausalVQA (Stage1/w), with MPE applied. As shown in Tab.~\ref{table_5}, multi-dimension training consistently outperforms single-dimension training in both settings. In the Stage1/wo case, exposure to diverse quality factors improves model performance across individual dimensions, indicating effective cross-task generalization. In Stage1/w, the combination of multi-dimension training and MPE further amplifies this advantage, resulting in greater accuracy gains compared to single-dimension training. These results highlight the synergistic benefits of heterogeneous training data and prompt diversity.
\begin{table*}[ht]
	\begin{center}
		\caption{Comparison of single-dimension and multi-dimension training strategies on the MVQA-test set. 
		}\label{table_5}
		\includegraphics[width=0.9\linewidth]{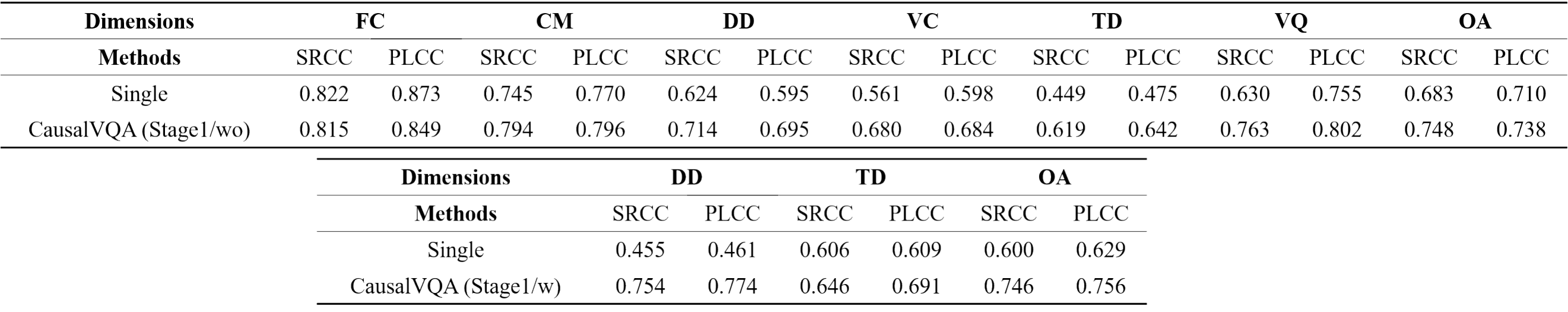}\\
	\end{center}
\end{table*}
\begin{table*}[ht]
	\begin{center}
		\caption{Ablation study showing the performance of general-purpose MLLMs before and after fine-tuning on the MVQA-68K dataset. 
		}\label{table_6}
		\includegraphics[width=0.9\linewidth]{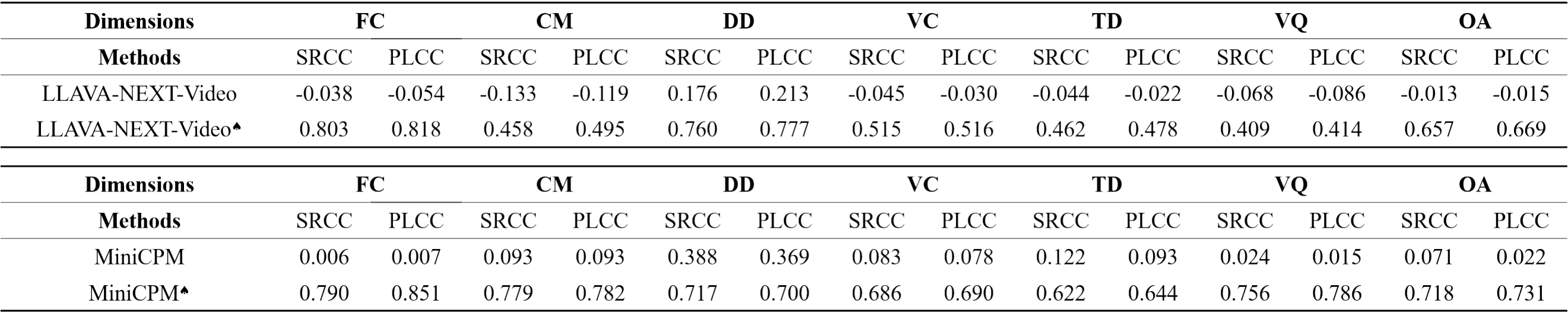}\\
	\end{center}
\end{table*}

\subsection{Ablation Study}
To assess the impact of MVQA-68K, we fine-tune several general-purpose MLLMs under a unified training setup. These models initially perform poorly on MVQA-test, with many dimension-wise correlations near zero or negative. After fine-tuning, as shown in Tab.~\ref{table_6}, all models (marked with $\spadesuit$) exhibit substantial gains in both SRCC and PLCC across all dimensions. Among them, MiniCPM shows the most significant improvement, with its SRCC on camera movement rising from 0.093 to 0.800 and aesthetics from 0.071 to 0.720. These results demonstrate that MVQA-68K effectively equips general MLLMs with the ability to model diverse and perceptual video quality factors.
\begin{table}[ht]
  \begin{center}
  \caption{Ablation results of LLaVA-Next-Video and MiniCPM on five public VQA benchmarks before and after fine-tuning on MVQA-68K. 
}\label{table_7}
\includegraphics[width=\linewidth]{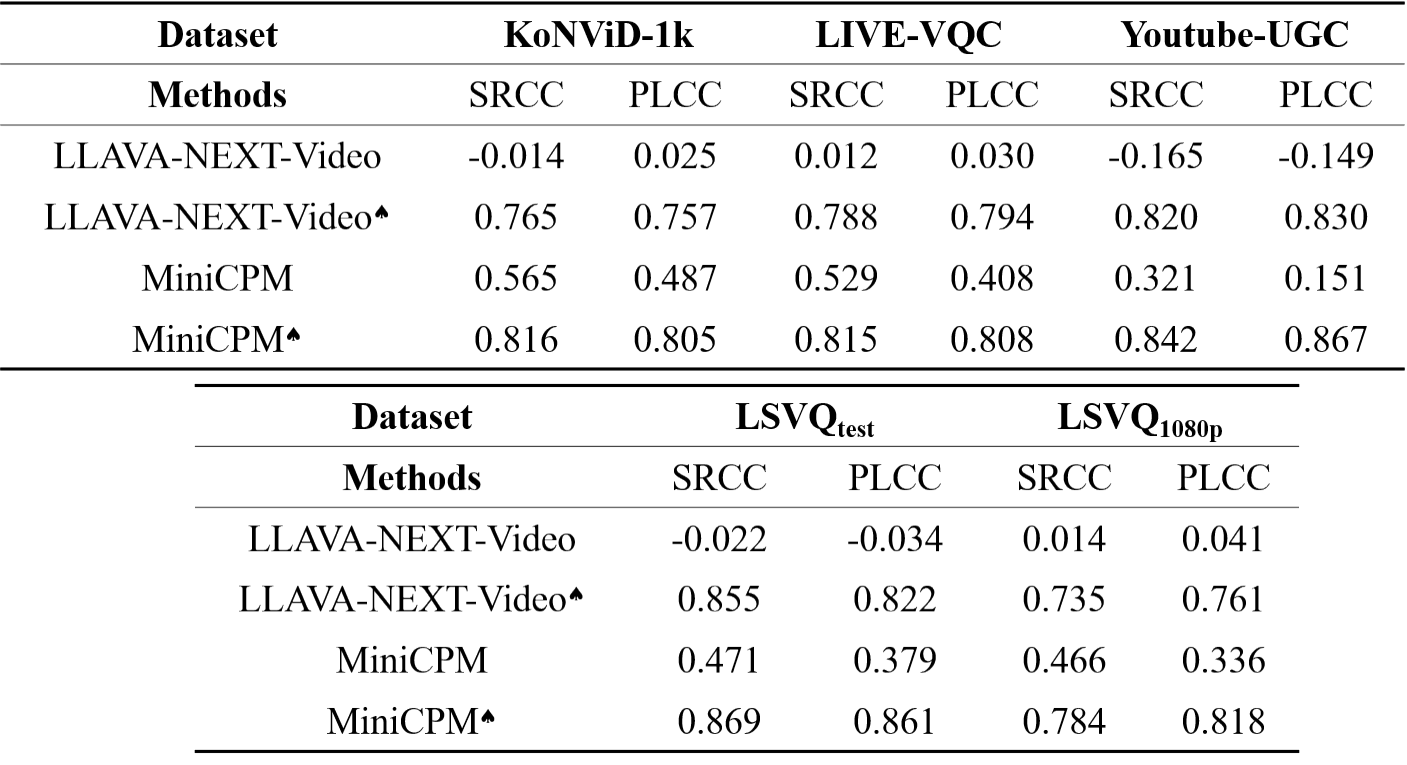}\\
  \end{center}
\end{table}

To evaluate the generalization benefit of MVQA-68K, we fine-tune LLaVA-Next-Video and MiniCPM on MVQA-68K and test them on five public VQA benchmarks. Both models initially perform poorly, with near-zero or negative correlations. After fine-tuning, as shown in Tab.~\ref{table_7}, they achieve substantial improvements across all datasets. LLaVA-Next-Video’s SRCC on YouTube-UGC increases from –0.165 to 0.820, while MiniCPM’s improves from 0.321 to 0.842. Similar gains are observed on KoNViD-1k, LIVE-VQC, LSVQ, and LSVQ-1080P.  

As shown in Tab.~\ref{table_8}, to isolate the impact of causal reasoning annotations on VQA performance, we train a dedicated variant called CausalVQA (Stage2-only). This model is trained using only the 7,217 question-answer pairs from the MVQA-68K dataset that are annotated with causal reasoning explanations. The training configuration and model architecture are kept identical to the other CausalVQA variants; the only difference is that the training data is restricted to this causal reasoning subset. This experiment evaluates whether this limited yet focused dataset slice can significantly enhance video quality assessment capabilities in MLLMs by infusing cause-effect understanding into the learning process. For comparison, we consider three previously evaluated CausalVQA models as baselines.
\begin{table}[ht]
  \begin{center}
  \caption{Performance of CausalVQA (Stage2-only) trained exclusively on 7,217 causal reasoning QA pairs, compared with other CausalVQA variants on the MVQA-test set. 
}\label{table_8}
\includegraphics[width=\linewidth]{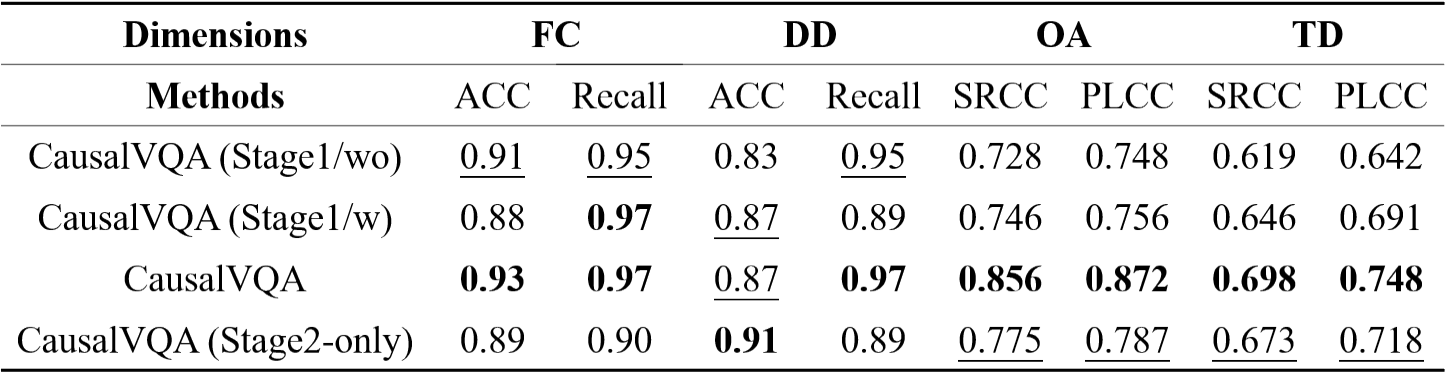}\\
  \end{center}
\end{table}

Although trained on only 11\% of the full dataset, the CausalVQA (Stage2-only) model achieves performance comparable to the Stage1/wo and Stage1/w baselines. Despite having access to far fewer training samples, it reaches similar accuracy and ranking correlations, and further narrows the performance gap with the full CausalVQA model. Notably, on subjective dimensions such as aesthetics and video composition, its results are only slightly lower than those of the full model. These findings demonstrate that causal reasoning annotations alone can significantly enhance the model’s ability to evaluate nuanced quality factors. This underscores the importance of incorporating cause-effect alignment  to improve interpretability and generalization in VQA.
\section{Conclusion}
In this paper, we presented \textbf{MVQA-68K}, a large-scale video quality assessment dataset containing over 68,000 annotated question–answer pairs across seven fine-grained quality dimensions. A key innovation of our dataset is the inclusion of structured causal rationale explanations, enabling models to explicitly reason about video quality in a human-aligned manner, addressing interpretability gaps in traditional VQA.

We further proposed the semantic-logit multi-prompt ensemble (\textbf{SemLogit-MPE}) strategy, which aggregates model predictions over semantically diverse prompts to enhance prediction consistency and stability. Leveraging these contributions, our resulting \textbf{CausalVQA} framework demonstrated superior performance over existing state-of-the-art models on both the MVQA-68K test set and multiple public VQA benchmarks. Ablation studies confirmed that even a small subset of causal annotations significantly improves model interpretability and generalization.

Overall, this work provides a robust foundation for developing interpretable, human-aligned video quality assessment systems using large multimodal language models.

\bibliographystyle{ACM-Reference-Format}
\balance
\bibliography{main}

%
%

\end{document}